# Machine Learning Based Source Code Classification Using Syntax Oriented Features

Shaul Zevin
Computer Industry
shaul.zevin@gmail.com

Catherine Holzem
Victoria University of Wellington
catherine.holzem@gmx.net

## ABSTRACT

As of today the programming language of the vast majority of the published source code is manually specified or programmatically assigned based on the sole file extension. In this paper we show that the source code programming language identification task can be fully automated using machine learning techniques. We first define the criteria that a production-level automatic programming language identification solution should meet. Our criteria include accuracy, programming language coverage, extensibility and performance. We then describe our approach: How training files are preprocessed for extracting features that mimic grammar productions, and then how these extracted 'grammar productions' are used for the training and testing of our classifier. We achieve a 99% accuracy rate while classifying 29 of the most popular programming languages with a Maximum Entropy classifier.

## 1. INTRODUCTION

The need for source-code-to-programming-language classification arises in use cases such as syntax highlighting, source code repositories indexing and estimation of trends in programming language popularity.

Popular syntax highlighters take different approaches: SyntaxHighlighter[1] and many other such tools ask the user to annotate the source code with the name of the matching programming language, which effectively boils down to manual programming language identification. Highlight.js[2] on the other hand uses the success-rate of the highlighting process to identify the language: Rare language constructs are given more weight while ubiquitous constructs such as common numbers have no weight at all; And there are also constructs defined as illegal that cause the parser to drop the highlighting attempt for a given language. Google Code Prettify[3] finally bypasses the programming identification hurdle altogether by generalizing the highlighting so it will work independently of the language.

The approach for the labelling of source code varies also when it comes to source code repositories: SourceForge[4] does not maintain the programming language on the source code file level, it merely stores the programming language of the project as a whole as manually specified by the project submitter. GitHub[5]'s approach is more sophisticated: It uses the Linguist[6] library, which applies following cascade of strategies to identify the programming language of the submitted source code:

1. Detect Emacs/Vim modelines

2. Looksfor a shebang "#!/…"

3. Check if file extension is unique for the language

4. Heuristics - are there any tell-tale syntax sequences that we can use to identify the language

5. Naïve Bayesian classification

Linguist stops when a single programming language candidate is left. Note that identification by unique file extension (strategy 3) has precedence over machine-learning-based classification (strategy 5).

The ideal of automatic computation goes back to Charles Babbage's wish to eliminate the risk of error in the production of printed mathematical tables. To make the automatic computation ideal practically applicable, one has to consider aspects such as use cases coverage, performance, maintenance and testing. We argue that the automatic solution for programming language identification should be evaluated based on following criteria:

1. Accuracy
   The accuracy of the results must be evaluated on a data set collected from as many different sources as possible. As stated by Brian W. Kernighan[7] "Computer programs can be written many different ways and still achieve the same effect … We have come to learn, however, that functionally equivalent programs can have extremely important stylistic differences". Note that absolute accuracy is unattainable because of the existence of polyglots – source code valid for more than a single programming language.

2. Popular programming languages support [8][9]
   Note that there are at least 20 programming languages with popularity rates exceeding 1%. And over 50 languages that pass the 0.1% mark.

3. Extensibility in a reasonable amount of effort and time
   Programming language popularity is dynamic and may exhibit fluctuations of a few percent in any one-year period. Moreover new programming languages emerge all the time e.g. Swift 2014.

4. Independence of file extension
   A file extension will not be available for snippets, short portions of code as posted on question and answer sites such as Stack Overflow [10]
   Moreover, even when available, file extensions can be ambiguous (.h .m .pl) or just plainly wrong.

5. Performance
   We define performance as the time required for the identification of the programming language of a single source file.
   Programming language identification will be embedded into real-time human-computer interactions and should not increase the response time of the whole process significantly.

Despite its many practical applications automated programming language identification has received moderate attention from research. We have summed up the various attempts known to us based on our evaluation criteria as described above in Table 1.

**Table 1**
**Published Research**

| Publication | Method | Features | Number of identified languages | Test data set size (number of files) | Accuracy % |
|---|---|---|---|---|---|
| Ugurel et al. [11] | Support Vector Machine | Alphabetical sequences of characters separated by non-alphabetical characters | 10 | 300 | 89.041 |
| Klein et al. [12] | Statistical analysis | Statistical features like percentage of lines with most common language keyword or brackets distribution | 25 | 625 | 48.0 |
| Khasnabish et al. [13] | Multinomial Naïve Bayes classifier | Keyword counts | 10 | 2392 | 93.48 |

| Kennedy van Dam et al. [14] | Modified Kneser-Ney discounting classifier | Unigrams, bigrams and trigrams of sequences of characters separated by white spaces | 20 | 4000 | 96.9 |
| This paper | Maximum Entropy classifier | Unigrams, bigrams and trigrams of sequences of lexicalized tokens | 29 | 147843 | 99.0 |

## 2. METHOD

One can hypothesize the best approach for identifying the programming language of a piece of source code to be the selection of the language whose grammar parses the code successfully. Such a method cannot be implemented in practice for a number of reasons: First, to obtain a grammar for each language we try to identify is hardly possible. For example "C++ is pretty much impossible to parse by merely writing up its grammar" – Terence Parr, the author of ANTLR [15]. Second, the maintenance cost of grammars will be extremely high as languages evolve and new languages emerge over time. Finally the need to run the source code against every available parser would inevitably negatively impact performance.

Our chosen approach is to automatically derive a single grammar from the training data that will later in the process be used for parsing every training and test file. The grammar includes the most representative productions for every language in the training set.

The ability to build such a common grammar automatically relies on the fact that programming languages share a common syntax structure, with building blocks such as keywords, identifiers, constants and statements. As Aho [17] p. 112 et al. states "In many programming languages, the following classes cover most or all of the tokens:

- One token for each keyword. The pattern for a keyword is the same as the keyword itself.
- Tokens for the operators, either individually or in classes such as the token comparison.
- One token representing all identifiers.
- One or more tokens representing constants, such as numbers and literal strings.
- Tokens for each punctuation symbol, such as left and right parentheses, comma, and semicolon."

Once built the grammar is run against the source code files from the training set: The most representative grammar productions for each language are at this stage of the process selected to be used as features by the language classifier.

Once the feature set has been defined the classifier can be trained. Classifier training is a process in which a weight is assigned to each (feature, language) pair. Features (grammar productions) found more frequently in the training set for a particular language than in the training files for other languages will be assigned a higher weight for that language.

Once the classifier has been trained we run it on our test set to measure its accuracy: Each test file is parsed and matched grammar productions are extracted and fed to the trained classifier, which outputs probabilities for each language represented in the training set. To estimate the classifier accuracy, we compare the language for which the classifier returned the highest probability to the actual language of the test file.

Note that the distribution of grammar productions is assumed to be very similar in the training and test sets. This is a key assumption for obtaining a high classification accuracy rate on the test set.

In the next subsections we provide a detailed description of the grammar construction, classifier training and classifier testing processes.

## 2.1 File Preprocessing

Any source code file, whether from the training or from the test set, first goes through some preliminary file preprocessing as detailed below.

**Diagram 1 File Preprocessing**

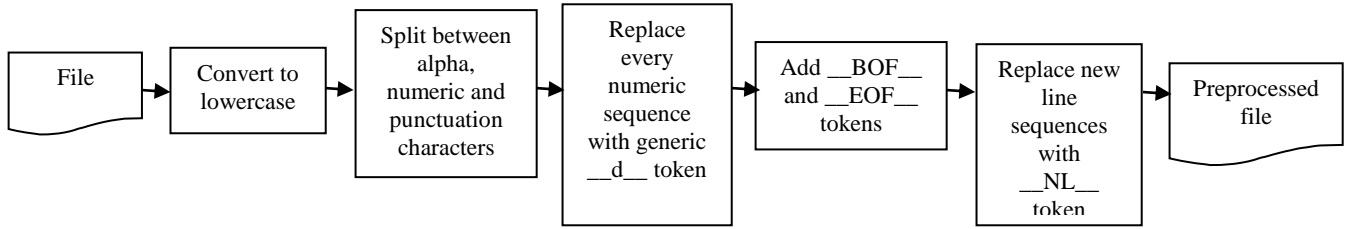

A conversion of all letters to lowercase is required since some programming languages (as e.g. Ada, Visual BASIC, GW Basic, Quite Basic, FORTRAN, SQL and Pascal) are case-insensitive.

Classical lexical analysis (see Appendix A) converts parsed source code into a stream of tokens. The lexer splits the source code into tokens by using white space characters as separators. Before applying the lexer we insert whitespaces between alpha, numeric and punctuation sequences as part of the file preprocessing since each such sequence represents a different type of token.

All numeric sequences are considered as constants and we replace each numeric sequence with a generic 'number' token __d__.

__BOF__ and __EOF__ (beginning and end of file) tokens are inserted at the beginning and at the end of file. The rationale behind this addition is that some statements like java package declaration can appear only at the beginning or the end of the file.

Consecutive new line characters ae replaced with a single __NL__ token since statements in languages like Basic, FORTRAN and Matlab are separated by a new line.

**Example 1. File Preprocessing**

source code fragment '*FUNCTION("123")*' will be split into '*function (" __d__ ")*'.

## 2.2 Grammar Construction

**Diagram 2 Grammar Construction**

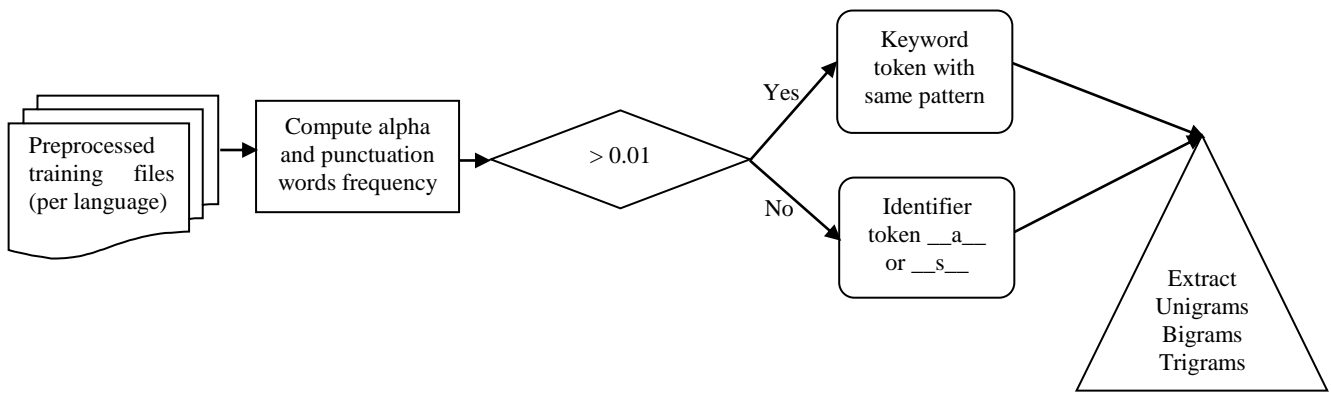

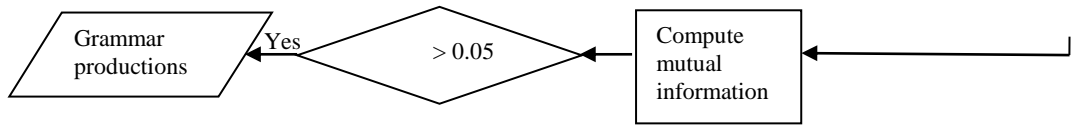

In a first step we process the set of training files available for each language in turn: We convert each word in the preprocessed training files (see section 2.1) to either the matching specific keyword token or to a generic 'identifier' token. To decide whether the word is an actual keyword or an identifier, we calculate the word frequency in the training files for the language being processed. The word frequency is expressed as the number of training files in which the word was found divided by total number of training files for the particular language. Words whose frequency exceeds 0.01 are considered as language keywords and replaced with a token of the same pattern i.e. the word "if" is replaced with a token "if". Words with a frequency below 0.01 are considered as identifiers. Alpha identifiers are replaced with the generic __a__ token and punctuation identifiers are replaced with __s__.

Note that, when calculating word frequencies, we deliberately skip language comments. Text such as copyright statement is not part of the language and should not be used for the language grammar extraction. The comments syntax is the only actual programming language syntax rule we have hardwired, all other grammar productions are derived from the training set in a next step as described below.

**Example 2. Keywords and identifiers replacement**

Preprocessed training file:

< ul class =" democrats "> __NL__
    < li > Clinton < / li > __NL__
</ ul > __NL__

Infrequent words *democrats* and *Clinton* are replaced with identifier token __a__, while frequent words such as *class* are replaced with a token of the same pattern:

< ul class =" __a__ "> __NL__
    < li > __a__ </ li > __NL__
</ ul > __NL__

In a second step we extract all unigrams, bigrams and trigrams from the token stream output of the first step. These extracted n-grams are the candidates for our grammar production selection process.

**Example 3. N-grams as extracted from Example 2.**
Unigrams: <, ul, class, =", __a__, ">, >, __NL__, li, </
Bigrams: < ul, ul class, class =", =" __a__, __a__ "> , "> __NL__, __NL__ <, < li, li >, > __a__, __a__ </, </ li, > __NL__, __NL__ </, </ ul, ul >
Trigrams: < ul class, ul class =", class =" __a__, =" __a__ ">, __a__ "> __NL__, "> __NL__ <, __NL__ < li, < li >, li > __a__, > __a__ </, __a__ </ li, </ li >, li > __NL__, > __NL__ </, __NL__ </ ul, </ ul >, ul > __NL__

The third and last step of our grammar construction process is the selection of the most informative grammar productions from the n-gram set generated during the second step. We use the MI (Mutual Information) index to determine which n-grams should be retained as grammar productions. The MI index measures how much information the presence/absence of a feature contributes to making the correct classification decision.

If *f* is a nominal feature with k different values and *l* is your target class with m classes, the MI of *f* is given by:

**Equation 1. Mutual Information**

$$MI = \sum_{i=1}^{k} \sum_{j=1}^{m} P(f_i, l_j) \log \frac{P(f_i, l_j)}{P(f_i) p(l_j)}$$

In our case k=2 and $f_1$=0/ $f_2$=1 denote absence/presence of a grammar production in a training file. Our m=29, the number of classified programming languages.

Only n-grams with an MI index above 0.05 make it into our grammar.

**Example 4. Grammar productions as selected from example 3 (MI above 0.05 threshold) followed by MI value**
*class* 0.297, *__NL__ <* 0.203, *__NL__ </* 0.189, *> __NL__ </* 0.180, *"> __NL__* 0.167, *"> __NL__ <* 0.163, *class ="* 0.143, *ul* 0.109, *_NL__ < li* 0.093, *__NL__ </ ul* 0.089, *__a__ "> __NL__* 0.071, *ul class* 0.0589, *ul class ="* 0.0584, *< ul class* 0.0576, *class =" __a__* 0.0546

## 2.3 Classifier training

We have opted for a maximum entropy (maxent) classifier. "The motivating idea behind maximum entropy is that one should prefer the most uniform models that also satisfy any given constraints." [16]. A nice property of the maxent classifier, as can be seen from the discussion below, is that no assumption is made on the relationships between features (our 'grammar productions'). For example an assumption on feature independency is not required. Since grammar productions are clearly dependent, the maxent classifier seemed like a promising classifier choice for our particular use case.

To project the programming language identification task onto the maximum entropy model, we define the following notation:

$L$ the set of all supported languages $l_j$

$S$ the set of all preprocessed training files

$l(s)$ the programming language of the training file $s$

$G$ the set of all grammar productions $g_i$ we have extracted in 2.2

$g_{i,j}(s,l)$ a grammar production indicator function from $S \times L$ to {0,1}. The function returns 1 if the sample $s$ contains grammar production $g_i$ and $l == l_j$. It returns 0 otherwise. The total number of such indicator functions is $|G| \times |L|$

$\lambda_{i,j}$ the weight of matched grammar production $i$ for language $j$. The values of these weights are computed during the classifier training process.

$p(l \mid s)$ a modeled conditional probability of language $l$ given sample $s$

$\tilde{p}(s)$ an empirical probability of a sample $s$

$H(p) = -\sum_{s \in S} \tilde{p}(s) p(l \mid s) \log p(l \mid s)$ is an entropy function of a modeled conditional probability

$LogLik(S) = \sum_{s \in S} \log(p(l(s) \mid s))$ is the log likelihood of $p(l \mid s)$ over the training set S

The constraints of our maximum entropy model are described in Equation 2. It requires each grammar production indicator function $g_{i,j}$ to result in model-predicted counts matching the empirical ones for the training set.

**Equation 2. Maximum entropy model constraints**

$$\forall g_i, l_j \quad \sum_{s \in S} p(l_j \mid s) g_{i,j}(s, l_j) = \sum_{s \in S} g_{i,j}(s, l(s))$$

It can be shown [20],[21] that the conditional probability distribution $p(l \mid s)$ that satisfies the constraints as described by Equation 2, and which has the form of Equation 3, will maximize the entropy $H(p)$. Furthermore such a function is unique.

**Equation 3. Probability of sample $s$ to be classified with language $l_j$**

$$p(l_j \mid s, \lambda) = \frac{\exp \sum_{g_i \in G} \lambda_{i,j} g_{i,j}(s, l_j)}{\sum_{l_k \in L} \exp \sum_{g_i \in G} \lambda_{i,k} g_{i,k}(s, l_k)}$$

It can be also shown [20],[21] that the conditional probability distribution $p(l \mid s)$ that meets the maximum entropy requirements (i.e. satisfies constraints of Equation 2 and has the form of Equation 3) will also have the maximum log likelihood over the training set $LogLik(S)$. The log likelihood $LogLik(S)$ is a convex function with a single maximum. Therefore any numerical optimization package can be used to find optimal grammar production weights $\lambda_{i,j}$ by exploring the log likelihood function space.

For the log likelihood value calculation we need to check if sample $s$ contains grammar production $g$. The grammar production checker procedure, described next, is used to answer that question.

### 2.4 Grammar Production Checker

The GrammarProductionChecker procedure outputs true if sample $s$ contains grammar production $g$ and false otherwise.

```
procedure GrammarProductionChecker(sample s, grammar production g)
    # s is preprocessed (see 2.1)
    s' = preprocess s
    # s' ≡ w_1 w_2 ... w_N
    N = number of words in s'
    # loop on every word boundary
    for x in 1..N
        if g is unigram and g matches w_x
            return true
        endif
        if g is bigram and g matches w_x w_{x+1}
            return true
        endif
        if g is trigram and g matches w_x w_{x+1} w_{x+2}
            return true
        endif
    end # words loop
    return false
end # procedure
```

### 2.5 Classifier testing

Once the classifier has been trained, ie once values have been computed for the $\lambda_{i,j}$ weights in Equation 3, we need to evaluate the accuracy of the trained classifier on a set of unseen test files.

Each test file is preprocessed and parsed against the grammar extracted from the training set to obtain the set of features or grammar productions present in the file. The probability of each programming language is calculated by using Equation 3. The classifier outputs the language with the highest probability as the language detected for the file.

Matching the output of the classifier against the actual labels of the test files we obtain precision, recall and F-measure values for each programming language in the test set.

$$precision\ of\ the\ language = \frac{number\ of\ test\ samples\ correctly\ identified\ with\ the\ language}{number\ of\ test\ samples\ labeled\ with\ the\ language}$$

$$recall\ of\ the\ language = \frac{number\ of\ test\ samples\ correctly\ identified\ with\ the\ language}{number\ of\ test\ samples\ identified\ with\ the\ language}$$

$$F \text{ measure of the language} = 2 * \frac{precision * recall}{precision + recall}$$

Precision and recall measure 2 different aspects of the accuracy of the trained classifier. The precision statistic (one statistic per supported language) gives the probability of the actual language being detected for any sample in the particular language. The recall statistic (also per language) gives the probability of the detection of the particular language to be a true positive, ie for that language to be the actual language of the classified sample. The F-measure combines these 2 complementary measures of a classifier's accuracy in one unique statistic.

## 3. EXPERIMENT DESIGN

### 3.1 Training and Test Data Selection

Similarly to [12],[13] and [14] we sourced our training and test data from GitHub [5]. Each GitHub source code repository typically consists of the files for a single software project.

We created a web crawler that randomly selects a subset of all the GitHub repositories annotated with a particular programming language until obtaining more than 5,000 training and 5,000 test files for each programming language to be supported by our classifier. Typically a few hundred repositories were needed to collect enough files for each programming language. Each selected repository was assigned in its entirety to either training or testing.
Almost every repository in our data set was created by a different programmer. Duplicate files were removed, and files smaller than 3 bytes or bigger than 240,000 bytes were skipped.

In total our entire dataset includes 293,319 source files collected from 16,134 distinct GitHub repositories. 145,476 files were used for training and 147,843 files for testing.

### 3.2 File Preprocessing

File preprocessing as described in section 2.1 was implemented by defining a lexical grammar using the ANTLR [15] software library.

### 3.3 Grammar Construction

We use a slightly modified version of the lexical grammar we implemented for the file preprocessing stage to compute the frequency of use of each alpha and punctuation word present in the training set of a particular language (see section 2.2). The lexical grammar is language-aware, so that it knows to skip comments for each language and only include words in the actual code. Then we apply an additional lexical analysis pass, in which each alpha and punctuation token in the language training set is converted to either the specific matching keyword or a generic identifier depending on the frequency of the word in the training set (the threshold for a word to be considered as a keyword for a particular language is 1 %, ie the word has to be present in at least 1% of all the training files selected for the language).

We use the WEKA v. 3.7.13 [18] software package for feature extraction and selection: We first apply the StringToWordVector WEKA filter with the NGramTokenizer procedure on the lexer output for each training file. This filter extracts all possible unigrams, bigrams and trigrams from the normalized token stream and adds them to the global grammar production set. This set is further filtered to merely retain the most informative grammar productions. This is achieved by applying the InfoGainAttributeEval WEKA filter, which calculates the Mutual Information index of each feature in the set (see Equation 1) and filters out any feature with a MI index lower than a specified threshold value ( in our case 0.05).

The final grammar consists of 47,147 productions, 5,327 of which are unigrams, 18,687 bigrams and 23,133 trigrams.

### 3.4 Classifier training and testing

We use the Stanford NLP toolkit [19] v. 3.5.2 implementation for the training and testing of our maximum entropy classifier. This implementation internally uses the L-BFGS quasi-Newton numeric method [22] to find the log likelihood $LogLik(S)$ maximum.

Regularization is a method by which model overfitting can be avoided. Regularization is implemented by the Stanford NLP Toolkit by subtracting $\frac{\sum \lambda_{i,j}^2}{2\sigma^2}$ from the log likelihood value, so that large weight values are penalized. We have set $\sigma = 10$.

### 3.5 Software implementation

Our source code classification system is available online under http://ec2-52-37-126-112.us-west-2.compute.amazonaws.com/falstaff/. Our implementation is based on the Spring Boot java framework, which provides an easy integration with other applications through web services.

The software is installed on a Amazon EC2 t2.small machine. This server has very moderate characteristics – 2.4 GHz processor and 2.0 GB RAM.

The performance of our programming language identification service on t2.small is 0.1 sec per source code file on average.

## 4. Results

We compared the performance of the Maximum Entropy and of the Naïve Bayesian classifiers for the task at hand in the initial stages of our research. The Maximum Entropy classifier out-performed the Naïve Bayesian classifier in accuracy by an average 2.5% (see Table 2).

**Table 2**
**Average Results per Classifier**

|  | Precision | Recall | F |
|---|---|---|---|
| Maximum Entropy | 0.991 | 0.99 | 0.99 |
| Naïve Bayesian | 0.965 | 0.967 | 0.965 |

With a resulting F score of 0.99 our method achieves a 2.1% improvement in accuracy on the best results published so far (see Table 1): Kennedy Van Dam [14] achieved an F-score of 0.969 while using a modified Kneser-Ney discounting classifier.

Our results (see Table 3) were verified on a very diverse and large test set containing 147843 files and covering the most popular 29 programming languages.

**Table 3**
**Maximum Entropy classifier results per programming language**

| Language | Precision | Recall | F |
|---|---|---|---|
| Ada | 0.999 | 1.0 | 1.0 |
| BatchFile | 0.992 | 0.967 | 0.98 |
| BourneShellScript | 0.978 | 0.977 | 0.978 |
| C/C++ | 0.985 | 0.995 | 0.99 |
| CSharp | 0.999 | 0.999 | 0.999 |
| COBOL | 0.996 | 0.973 | 0.984 |
| CascadingStyleSheets | 0.993 | 0.99 | 0.991 |
| FORTRAN | 0.977 | 0.997 | 0.987 |
| Go | 0.996 | 0.996 | 0.996 |
| HTML | 0.976 | 0.976 | 0.976 |
| Haskell | 0.995 | 0.989 | 0.992 |
| Java | 0.999 | 1.0 | 1.0 |
| JavaScript | 0.966 | 0.987 | 0.976 |
| LISP | 0.997 | 0.991 | 0.994 |
| LaTeX | 0.986 | 0.992 | 0.989 |
| MATLABScriptFile | 0.992 | 0.995 | 0.994 |
| Objective-C | 0.994 | 0.98 | 0.987 |
| PHP | 0.998 | 0.988 | 0.993 |
| Pascal | 0.998 | 0.999 | 0.999 |
| Perl | 0.99 | 0.995 | 0.992 |
| Prolog | 0.988 | 0.985 | 0.986 |
| Python | 0.992 | 0.982 | 0.987 |
| R | 0.997 | 0.99 | 0.993 |
| Ruby | 0.987 | 0.995 | 0.991 |
| SQL | 0.994 | 0.987 | 0.99 |
| Scala | 0.995 | 0.997 | 0.997 |
| Swift | 0.993 | 0.997 | 0.995 |
| Tcl | 0.983 | 0.991 | 0.987 |
| VisualBasic | 0.999 | 1.0 | 1.0 |
| Average | 0.991 | 0.99 | 0.99 |

**Table 4**
**Confusion Table**

| | Ada | BatchFile | BourneShellScript | C/C++ | CSharp | COBOL | CascadingStyleSheets | FORTRAN | Go | HTML | Haskell | Java | JavaScript | LISP | LaTeX | MATLABScriptFile | Objective-C | PHP | Pascal | Perl | Prolog | Python | R | Ruby | SQL | Scala | Swift | Tcl | VisualBasic |
|---|---|---|---|---|---|---|---|---|---|---|---|---|---|---|---|---|---|---|---|---|---|---|---|---|---|---|---|---|---|
| Ada | 4999 | 0 | 0 | 0 | 0 | 0 | 0 | 0 | 1 | 0 | 0 | 0 | 0 | 0 | 0 | 0 | 0 | 0 | 0 | 0 | 0 | 0 | 0 | 0 | 2 | 0 | 0 | 0 | 0 |
| BatchFile | 0 | 4927 | 16 | 1 | 0 | 1 | 0 | 0 | 1 | 6 | 1 | 0 | 1 | 0 | 2 | 0 | 0 | 0 | 0 | 1 | 1 | 1 | 2 | 0 | 3 | 1 | 1 | 1 | 0 |
| BourneShellScript | 0 | 25 | 4879 | 3 | 0 | 0 | 0 | 0 | 0 | 2 | 1 | 0 | 12 | 0 | 2 | 3 | 0 | 0 | 0 | 4 | 2 | 25 | 15 | 2 | 3 | 0 | 0 | 9 | 0 |
| C/C++ | 1 | 3 | 4 | 10070 | 0 | 4 | 5 | 0 | 4 | 5 | 0 | 0 | 1 | 0 | 1 | 0 | 103 | 0 | 0 | 0 | 5 | 4 | 0 | 1 | 6 | 1 | 0 | 9 | 0 |
| CSharp | 0 | 0 | 1 | 0 | 5004 | 1 | 0 | 0 | 0 | 1 | 0 | 0 | 0 | 1 | 0 | 0 | 0 | 0 | 0 | 0 | 0 | 0 | 0 | 0 | 1 | 0 | 0 | 0 | 0 |
| COBOL | 0 | 3 | 1 | 0 | 0 | 1259 | 0 | 0 | 0 | 0 | 0 | 0 | 0 | 0 | 0 | 0 | 0 | 0 | 0 | 0 | 0 | 0 | 0 | 0 | 1 | 0 | 0 | 0 | 0 |
| CascadingStyleSheets | 0 | 3 | 13 | 0 | 0 | 0 | 5068 | 0 | 0 | 7 | 0 | 0 | 3 | 0 | 0 | 0 | 0 | 0 | 2 | 0 | 1 | 6 | 0 | 2 | 0 | 0 | 1 | 0 | 0 |
| FORTRAN | 0 | 12 | 0 | 6 | 0 | 0 | 1 | 4921 | 0 | 0 | 43 | 0 | 1 | 43 | 1 | 0 | 0 | 0 | 0 | 0 | 5 | 0 | 0 | 1 | 0 | 1 | 0 | 0 | 0 |
| Go | 1 | 0 | 4 | 1 | 0 | 0 | 1 | 0 | 5056 | 10 | 0 | 0 | 1 | 0 | 0 | 0 | 0 | 0 | 0 | 0 | 0 | 0 | 0 | 0 | 0 | 0 | 0 | 0 | 0 |
| HTML | 0 | 8 | 6 | 5 | 0 | 0 | 13 | 0 | 1 | 4889 | 0 | 0 | 21 | 0 | 1 | 1 | 1 | 54 | 1 | 0 | 0 | 1 | 4 | 1 | 0 | 0 | 0 | 0 | 0 |
| Haskell | 0 | 2 | 2 | 0 | 0 | 0 | 0 | 0 | 0 | 0 | 5051 | 0 | 2 | 0 | 0 | 1 | 0 | 0 | 0 | 3 | 12 | 2 | 0 | 1 | 0 | 1 | 0 | 0 | 0 |
| Java | 0 | 0 | 0 | 1 | 2 | 0 | 0 | 0 | 0 | 0 | 0 | 5061 | 0 | 0 | 0 | 0 | 0 | 0 | 0 | 0 | 0 | 0 | 0 | 0 | 0 | 1 | 0 | 0 | 0 |
| JavaScript | 0 | 38 | 1 | 1 | 1 | 0 | 26 | 0 | 1 | 37 | 3 | 0 | 4907 | 0 | 3 | 7 | 0 | 2 | 0 | 4 | 11 | 8 | 6 | 4 | 16 | 0 | 4 | 2 | 0 |
| LISP | 0 | 3 | 2 | 0 | 1 | 0 | 0 | 0 | 0 | 1 | 3 | 0 | 0 | 5019 | 1 | 0 | 0 | 0 | 0 | 0 | 0 | 1 | 1 | 0 | 0 | 0 | 0 | 0 | 0 |
| LaTeX | 0 | 5 | 4 | 2 | 0 | 3 | 0 | 5 | 1 | 19 | 4 | 0 | 1 | 3 | 5166 | 3 | 0 | 0 | 0 | 15 | 2 | 1 | 2 | 1 | 0 | 1 | 0 | 0 | 0 |
| MATLABScriptFile | 0 | 8 | 0 | 2 | 0 | 1 | 0 | 0 | 0 | 7 | 0 | 0 | 0 | 0 | 6 | 5056 | 0 | 0 | 0 | 1 | 9 | 0 | 4 | 1 | 0 | 0 | 0 | 0 | 1 |
| Objective-C | 0 | 0 | 0 | 23 | 0 | 0 | 1 | 0 | 0 | 1 | 1 | 0 | 1 | 0 | 0 | 0 | 5054 | 0 | 0 | 0 | 0 | 0 | 0 | 0 | 1 | 0 | 0 | 1 | 0 |
| PHP | 0 | 0 | 0 | 0 | 0 | 0 | 0 | 0 | 0 | 5 | 0 | 0 | 0 | 0 | 0 | 0 | 0 | 4949 | 0 | 2 | 2 | 0 | 0 | 1 | 2 | 0 | 0 | 0 | 0 |
| Pascal | 0 | 0 | 0 | 0 | 0 | 0 | 5 | 0 | 0 | 0 | 0 | 0 | 0 | 0 | 1 | 0 | 0 | 0 | 5006 | 1 | 0 | 0 | 0 | 0 | 1 | 0 | 0 | 0 | 0 |
| Perl | 0 | 1 | 8 | 2 | 0 | 0 | 0 | 0 | 0 | 6 | 0 | 0 | 5 | 0 | 3 | 2 | 0 | 0 | 0 | 4997 | 3 | 4 | 1 | 2 | 15 | 0 | 0 | 1 | 0 |
| Prolog | 0 | 13 | 1 | 0 | 0 | 12 | 1 | 10 | 0 | 0 | 1 | 0 | 0 | 1 | 11 | 2 | 0 | 0 | 0 | 0 | 4238 | 0 | 0 | 0 | 1 | 0 | 0 | 0 | 0 |
| Python | 0 | 0 | 10 | 2 | 0 | 3 | 0 | 0 | 0 | 2 | 1 | 0 | 3 | 0 | 1 | 2 | 1 | 0 | 1 | 0 | 1 | 4979 | 5 | 5 | 0 | 0 | 0 | 2 | 0 |
| R | 0 | 2 | 4 | 0 | 0 | 0 | 0 | 0 | 0 | 1 | 0 | 0 | 0 | 0 | 3 | 0 | 0 | 0 | 0 | 0 | 0 | 2 | 5026 | 2 | 0 | 0 | 1 | 0 | 0 |
| Ruby | 0 | 6 | 5 | 2 | 0 | 0 | 1 | 1 | 0 | 4 | 0 | 0 | 3 | 0 | 2 | 0 | 0 | 1 | 2 | 2 | 1 | 16 | 8 | 4968 | 1 | 2 | 0 | 8 | 0 |
| SQL | 0 | 3 | 8 | 0 | 0 | 0 | 3 | 0 | 0 | 2 | 0 | 0 | 1 | 0 | 0 | 0 | 0 | 2 | 0 | 3 | 0 | 3 | 0 | 0 | 5817 | 0 | 0 | 11 | 0 |
| Scala | 0 | 1 | 0 | 0 | 0 | 0 | 1 | 0 | 6 | 0 | 0 | 0 | 2 | 0 | 3 | 0 | 0 | 0 | 0 | 2 | 0 | 0 | 0 | 2 | 2 | 5030 | 5 | 0 | 0 |
| Swift | 0 | 2 | 0 | 0 | 0 | 2 | 0 | 0 | 6 | 2 | 0 | 1 | 4 | 0 | 1 | 0 | 1 | 0 | 0 | 0 | 13 | 2 | 0 | 0 | 2 | 5057 | 0 | 0 | 0 |
| Tcl | 0 | 32 | 24 | 1 | 2 | 0 | 2 | 0 | 0 | 0 | 0 | 0 | 4 | 0 | 1 | 0 | 0 | 0 | 3 | 0 | 1 | 1 | 3 | 13 | 0 | 0 | 4948 | 0 | 0 |
| VisualBasic | 0 | 0 | 0 | 0 | 0 | 2 | 0 | 0 | 0 | 0 | 0 | 0 | 0 | 0 | 0 | 0 | 0 | 0 | 0 | 0 | 1 | 0 | 1 | 0 | 0 | 0 | 0 | 0 | 5043 |

## 4.1. Results discussion

Results vary from programming language to programming language (see Table 3).

We speculate the main contributing factors for misclassification (see Table 4) to be:

- Short polyglots.
  Some files are syntactically correct and equally probable in more than one language. Typically such files consist of a few lines of code.
  C/C++ and Objectice C header files provides a good example for this use case.
  Another example is given by Tcl commands that also exist in the Unix shell.

- Bidirectional embedding of one programming language into another.
  Classification seems to cope fairly well with unidirectional embedding like SQL embedding into other languages. Languages that can be embedded into each other are more challenging for the classifier.
  A good example is HTML and PHP.
  Another example is JavaScript and HTML.

# 5. Conclusions

Our method shows that source code to programming language classification can be done in accordance with the criteria we set out to define for a production-ready implementation in the Introduction section:

1. Method achieves F=0.99 accuracy measured on 147843 source files collected from diverse sources.

2. Method supports most popular 29 programming languages.

3. Method is fully automated and does not require any knowledge of the programming languages it identifies. (Except for the grammar construction where we have used comments syntax knowledge)

4. Method does not rely on programming language file extension or any other file metadata.

5. Method is implemented with an average 0.1 sec per identification performance on a very modest server configuration.

Programming language grammatical rules have a recursive nature. In the future we would like to explore the possibility of using deep learning Recurrent Neural Networks to improve our results even further.

## Appendix A. Programming Language Grammar Definitions

Definitions below are quotes from the classical book "Compilers: Principles, Techniques and Tools" 2[nd] edition written by Aho, A., Lam, M., Sethi, R., Ullman, J., Cooper, K., Torczon, L., & Muchnick, S [17]

*Context Free Grammar* [p. 42] - A context-free grammar has four components:

1. A set of terminal symbols, sometimes referred to as "tokens." The terminals are the elementary symbols of the language defined by the grammar.
2. A set of nonterminals, sometimes called "syntactic variables." Each nonterminal represents a set of strings of terminals, in a manner we shall describe.
3. A set of productions, where each production consists of a nonterminal called the head or left side of the production, an arrow, and a sequence of terminals and/or nonterminals , called the body or right side of the production. The intuitive intent of a production is to specify one of the written forrms of a construct; if the head nonterminal represents a construct, then the body represents a written form of the construct.
4. A designation of one of the nonterminals as the start symbol.

*Lexical Analyzer* [p. 43] - In a compiler, the lexical analyzer reads the characters of the source program,

groups them into lexically meaningful units called lexemes, and produces as output tokens presenting these lexemes. A token consists of two components, a token name and an attribute value. The token names are abstract symbols that are used by the parser for syntax analysis. Often, we shall call these token names terminals, since they appear as terminal symbols in the grammar for a programming language.

## 6. ACKNOWLEDGMENTS

## 7. REFERENCES

[1] SyntaxHighlighter http://alexgorbatchev.com/SyntaxHighlighter/.

[2] Highlight.js https://highlightjs.org/.

[3] Google Code Prettify https://github.com/google/code-prettify.

[4] SourceForge https://sourceforge.net/.

[5] GitHub https://github.com/.

[6] Linguist https://github.com/github/linguist.

[7] Kernighan, Brian W., and Phillip J. Plauger. "Programming style: Examples and counterexamples." *ACM Computing Surveys (CSUR)* 6.4 (1974): 303-319.

[8] TIOBE index http://www.tiobe.com/tiobe-index/.

[9] PYPL index http://pypl.github.io/PYPL.html.

[10] Stack Overflow http://stackoverflow.com/

[11] Ugurel, Secil, Robert Krovetz, and C. Lee Giles. "What's the code?: automatic classification of source code archives." *Proceedings of the eighth ACM SIGKDD international conference on Knowledge discovery and data mining*. ACM, 2002.

[12] Klein, David, Kyle Murray, and Simon Weber. "Algorithmic programming language identification." *arXiv preprint arXiv:1106.4064* (2011).

[13] Khasnabish, Jyotiska Nath, et al. "Detecting Programming Language from Source Code Using Bayesian Learning Techniques." *International Workshop on Machine Learning and Data Mining in Pattern Recognition*. Springer International Publishing, 2014.

[14] van Dam, Juriaan Kennedy, and Vadim Zaytsev. "Software Language Identification with Natural Language Classifiers." *2016 IEEE 23rd International Conference on Software Analysis, Evolution, and Reengineering (SANER)*. Vol. 1. IEEE, 2016.

[15] Parr, Terence. *The definitive ANTLR 4 reference*. Pragmatic Bookshelf, 2013.

[16] Nigam, Kamal, John Lafferty, and Andrew McCallum. "Using maximum entropy for text classification." *IJCAI-99 workshop on machine learning for information filtering*. Vol. 1. 1999.

[17] Aho, A., Lam, M., Sethi, R., Ullman, J., Cooper, K., Torczon, L., & Muchnick, S. (2007). Compilers: Principles, Techniques and Tools.

[18] Witten, Ian H., and Eibe Frank. *Data Mining: Practical machine learning tools and techniques*. Morgan Kaufmann, 2005.

[19] Manning, Christopher D., et al. "The Stanford CoreNLP Natural Language Processing Toolkit." *ACL (System Demonstrations)*. 2014.

[20] Ratnaparkhi, Adwait. "A simple introduction to maximum entropy models for natural language processing." *IRCS Technical Reports Series* (1997): 81.

[21] Della Pietra, Stephen, Vincent Della Pietra, and John Lafferty. "Inducing features of random fields." *IEEE transactions on pattern analysis and machine intelligence* 19.4 (1997): 380-393.

[22] Nocedal, Jorge, and Stephen Wright. *Numerical optimization*. Springer Science & Business Media, 2006.